\documentclass[pdflatex,sn-mathphys-num]{sn-jnl}
\usepackage[utf8]{inputenc}

\usepackage{graphicx}%
\usepackage{multirow}%
\usepackage{amsmath,amssymb,amsfonts}%
\usepackage{amsthm}%
\usepackage[title]{appendix}%
\usepackage{xcolor}%
\usepackage{textcomp}%
\usepackage{manyfoot}%
\usepackage{booktabs}%
\usepackage{tabularx}
\usepackage{float}
\usepackage{placeins}
\usepackage{svg}
\usepackage{algorithm}%
\usepackage{algorithmicx}%
\usepackage{algpseudocode}%
\usepackage{listings}%

\usepackage{array}
\usepackage{booktabs}
\usepackage{multirow}
\usepackage{makecell}
\usepackage{graphicx}
\usepackage{adjustbox}

\usepackage{pifont}  
\usepackage{xcolor}  
\usepackage{booktabs}  

\usepackage{float}
\begin{document}

\title[COMPASS: The explainable agentic framework for Sovereignty, Sustainability, Compliance, and Ethics]{COMPASS: The explainable agentic framework for Sovereignty, Sustainability, Compliance, and Ethics}

\author*[1]{\fnm{Jean-Sébastien} \sur{Dessureault}}\email{jean-sebastien.dessureault@uqtr.ca}
\author[2]{\fnm{Alain-Thierry} \sur{Iliho Manzi}}\email{Alain.Thierry.Iliho.Manzi@uqtr.ca}
\author[2]{\fnm{Soukaina} \sur{Alaoui Ismaili}}\email{Soukaina.Alaoui.Ismaili@uqtr.ca}
\author[2]{\fnm{Khadim} \sur{Lo}}\email{Khadim.Lo@uqtr.ca}
\author[2]{\fnm{Mireille} \sur{Lalancette}}\email{Mireille.Lalancette@uqtr.ca}
\author[3]{\fnm{Éric} \sur{Bélanger}}\email{eric.belanger3@mcgill.ca}

\affil*[1]{\orgdiv{Mathematics and Computer science}, \orgname{Université du Québec à Trois-Rivières}, \orgaddress{\street{3351 Boul. des Forges}, \city{Trois-Rivières}, \postcode{G8Z 4M3}, \state{Québec}, \country{Canada}}}

\affil*[2]{\orgdiv{LSSI - Laboratory of Signal and System Integration, Electrical and Computer Engineering Department}, \orgname{Université du Québec à Trois-Rivières}, \orgaddress{\street{3351 Boul. des Forges}, \city{Trois-Rivières}, \postcode{G8Z 4M3}, \state{Québec}, \country{Canada}}}

\affil[2]{\orgdiv{Social Communication Department}, \orgname{Université du Québec à Trois-Rivières}, \orgaddress{\street{3351 Boul. des Forges}, \city{Trois-Rivières}, \postcode{G8Z 4M3}, \state{Québec}, \country{Canada}}}

\affil[3]{\orgdiv{Department of Political Science}, \orgname{McGill University}, \orgaddress{\street{845 rue Sherbrooke}, \city{Montréal}, \postcode{H3A 0G4}, \state{Québec}, \country{Canada}}}

\abstract{The rapid proliferation of large language model (LLM)-based agentic systems raises critical concerns regarding digital sovereignty, environmental sustainability, regulatory compliance, and ethical alignment. Whilst existing frameworks address individual dimensions in isolation, no unified architecture systematically integrates these imperatives into the decision-making processes of autonomous agents. This paper introduces the COMPASS (Compliance and Orchestration for 
Multi-dimensional Principles in Autonomous Systems with Sovereignty) Framework, a novel multi-agent orchestration system designed to enforce value-aligned AI through modular, extensible governance mechanisms. The framework comprises an  Orchestrator and four specialised sub-agents—addressing sovereignty, carbon-aware computing, compliance, and ethics, each augmented with Retrieval-Augmented Generation (RAG) to ground evaluations in verified, context-specific documents. By employing an LLM-as-a-judge methodology, the system assigns quantitative scores and generates explainable justifications for each assessment dimension, enabling real-time arbitration of conflicting objectives. We validate the architecture through automated evaluation, demonstrating that RAG integration significantly enhances semantic coherence (as measured by BERTScore) and mitigates the hallucination risks inherent in non-augmented models. Our results indicate that the framework's composition-based design facilitates seamless integration into diverse application domains whilst preserving interpretability and traceability. Although current limitations include the absence of human-in-the-loop validation and underdeveloped action-selection capabilities, the Framework provides a robust methodological foundation for the responsible deployment of AI. Future work will incorporate hybrid validation protocols, optimised document curation strategies, and code-level ethical analysis to further strengthen the system's reliability and societal acceptability.}


\maketitle

\section{Introduction}\label{Introduction}

LLM-based agentic AI is becoming increasingly ubiquitous in our daily lives; consequently, we must plan its deployment to ensure that it remains socially acceptable. However, social acceptability is a complex, multi-dimensional concept that varies significantly across global regions and cultural values. Key dimensions include digital sovereignty, energy frugality, regulatory compliance, and, inevitably, ethics. These dimensions frequently conflict, making the identification of an optimal agentic solution highly challenging. For instance, the most energy-efficient model may be inferior in terms of digital sovereignty, or vice versa.
This paper proposes a novel LLM-based agentic framework that analyses these dimensions and offers concrete optimisation solutions tailored to the deployment region's specific values. Specifically, the framework utilises a multi-agent architecture anchored by a core "Synchronising Agent" that is designed to be inherited by other agentic object structures. Thus, any agent, regardless of its application domain, can inherit the knowledge and methods required to optimise its operation across the four aforementioned dimensions, thereby fostering its social acceptability.

\subsection{Agentic AI}
Autonomous language agents (or generative agents, per \cite{sumers_cognitive_2024}) are an emerging class of systems that integrate natural language understanding with decision-making capabilities to execute complex tasks in dynamic environments. Powered by pre-trained LLMs, these agents leverage their core reasoning engines to interact with digital tools and real-world data, enabling autonomous reasoning, planning, and action. 

As outlined by \cite{sumers_cognitive_2024}, such agents operate within a cognitive architecture where the LLM interprets user goals, retrieves relevant knowledge, and performs actions—either internally (through reasoning and planning) or externally (via tool use or communication). This framework integrates classical principles of cognitive systems with the generalisation capabilities of foundation models.

As \cite{huang_understanding_2024} argues, LLM-based agents transcend the limitations of static, prompt-augmented language models by functioning as dynamic, structured planning systems. Their proposed typology, which includes prompt-based agents, tool-augmented agents, and memory-augmented agents, captures varying degrees of autonomy and environmental interaction.

Meanwhile, the CoALA framework \cite{sumers_cognitive_2024} anchors the design of language agents in principles of cognitive science. It introduces a modular architecture that distinguishes between memory systems (semantic, episodic, and procedural), action modalities (internal and external), and long-term planning mechanisms, thereby integrating traditional AI planning with the adaptive reasoning capabilities of LLMs.

Another notable approach is AutoGPT \cite{significant_gravitas_autogpt_2025}, which demonstrates how LLM agents can autonomously decompose high-level goals into subtasks, execute them, and iteratively refine their plans through a planning-act-revise loop. While less formally structured than frameworks such as ReAct \cite{yao2022react} or CoALA \cite{sumers_cognitive_2024}, AutoGPT demonstrates the potential for open-ended, language-driven autonomous planning. Recent surveys \cite{huynh_large_2025} \cite{acharya_agentic_2025} and reviews \cite{hosseini_role_2025} outline the latest advancements in agentic AI and its future projections.

\subsection{Digital Sovereignty}
Digital sovereignty \cite{floridi_fight_2020} \cite{pohle_digital_2020} addresses the critical challenge of technological dependence and the erosion of autonomy in an increasingly centralized digital ecosystem. The fundamental problem lies in the asymmetric power dynamic between local entities and global hyperscalers; in this context, organizations often lack control over the infrastructure, data storage, and algorithmic processes essential to their operations. This reliance exposes actors to geopolitical risks, legal vulnerabilities regarding extraterritorial data laws, and the opacity of "black box" proprietary systems. Consequently, the absence of a sovereign framework limits the capacity of nations and industries to self-determine their digital future, compelling them to operate under constraints imposed by foreign technology providers. Several initiatives, such as the European Union's Gaia-X \footnote{Europe's Gaia-X: \url{https://gaia-x.eu/}}, aim to strengthen digital sovereignty. Similarly, Canada has introduced a framework \footnote{Canadian framework: \url{https://www.canada.ca/en/government/system/digital-government/digital-government-innovations/cloud-services/digital-sovereignty/digital-sovereignty-framework-improve-digital-readiness.html}} specifically designed to counteract this digital dependence. 

\subsection{Carbon-Aware Computing}
The environmental impact of LLMs is a critical area of investigation in sustainable computing. While the substantial carbon footprint associated with the pre-training phase of foundation models is well documented, recent scholarship emphasizes the growing significance of inference-related emissions. This operational aspect is particularly acute in the context of agentic AI, where iterative prompting and complex reasoning tasks increase energy consumption per task exponentially. Furthermore, a comprehensive Life Cycle Assessment necessitates accounting for the embodied carbon in hardware manufacturing and the intensive energy demands of data centres, highlighting the urgent need for carbon-efficient model architectures and rigorous Greenhouse Gas monitoring methodologies.

This issue is currently being addressed through a diverse array of academic and private initiatives, such as the \textit{Green AI Institute} \footnote{Green IA Institute: \url{https://www.greenai.institute/home}} and the Canadian firm \textit{GenerIA} \footnote{GenerIA: \url{https://generia.ai/en/home}}, alongside significant contributions in recent literature \cite{cowls_ai_2023} \cite{toumi_ai_2025}. Furthermore, several software packages designed for energy management, optimisation, and awareness have become available. Among these, \textit{CodeCarbon} \footnote{CodeCarbon: \url{https://mlco2.github.io/codecarbon/index.html}} is particularly noteworthy, having been integrated into the architecture of emerging machine learning frameworks \cite{dessureault_ai2_2023-journal}. A comparable alternative for carbon tracking is the \textit{eco2AI} library \cite{budennyy_eco2ai_2022}.

\subsection{AI compliance}
The alignment of artificial intelligence systems with national legislative frameworks has emerged as a paramount operational requirement for system architects. While the European Union generally sets the global precedent for regulation, the legal heterogeneity across borders necessitates a strictly localised approach to compliance. In the Canadian context, this transition from abstract ethical guidelines to enforceable statutes is exemplified by the \textit{Artificial Intelligence and Data Act} (AIDA), introduced under Bill C-27 \cite{parl_c27_2022}. As noted by \cite{scassa2023aida}, the Canadian regulatory approach contrasts with European models by prioritising agile frameworks tailored specifically to high-impact systems, thereby demanding rigorous adherence to domestic sovereignty regarding data governance and liability. Consequently, regulatory compliance cannot remain a mere post-deployment audit function; rather, it must be integrated as a foundational element of the architectural design to effectively navigate these evolving regulatory markets \cite{hadfield2023regulatory}.

\subsection{AI ethics}
Beyond strict statutory adherence, the ethical dimension of AI mandates a transition from abstract principles to operationalisable technical standards. The challenge for system architects lies in translating qualitative values—such as fairness, autonomy, and privacy—into quantitative objective functions that autonomous agents cannot violate. As emphasised by the \textit{Montreal Declaration for Responsible AI}, the development of algorithmic systems must prioritise the well-being of all sentient beings while actively mitigating the risks of algorithmic bias and opacity \cite{montreal2018declaration}. Consequently, ethical governance is no longer a peripheral oversight mechanism but a core component of the "Responsible AI" framework, requiring continuous monitoring of inference fairness and interpretability throughout the system's lifecycle \cite{dignum2019responsible}.

Assessing the ethical aspects of LLMs and LLM-based agents in real-time remains a complex challenge. However, emerging research begins to address this issue. For instance, \cite{park_generative_2023} introduces a simulation of autonomous agents within a virtual village, where the behaviours of virtual citizens are observed and analysed. Similarly, \cite{Dessureault_Sandbox_2025} presents a simulation wherein agents act as heads of state leading a nation. Their interactions, exchanges, thoughts, and decisions are analysed through both communicational and ethical lenses. Methodologies employing the LLM-as-a-judge technique \cite{liu_agentbench_2023} \cite{zheng2024judging} are increasingly proposed, incorporating specific reflection on the ethical dimension. It is crucial to recognise that agents are rapidly evolving; consequently, ethical risks must be assessed not only for the present context but also for the near future, as Human-Level AI (AGI) and even superintelligence loom on the horizon. These domains of a potentially imminent future are already being documented \cite{bostrom2014superintelligence} \cite{dessureault_ethics_2025}.

\subsection{LLM-as-Judge Methodologies}

Recent work employs LLMs to evaluate the outputs of other models. 
\textbf{MT-Bench} \cite{zheng2023judging} and \textbf{AlpacaEval} 
\cite{alpaca_eval} use GPT-4 as an automated judge to assess generation quality. \cite{chiang2024llm_judge} demonstrates correlation with human judgment for reasoning tasks. However, existing judge frameworks focus on \textit{quality assessment} rather than on multi-dimensional normative evaluation that simultaneously encompasses sovereignty, sustainability, compliance, and ethics.



\subsection{Gap Analysis, Positioning and Contribution}

Table \ref{tab:related_work} summarises how existing approaches address the four dimensions central to our work. While individual solutions excel in specific areas, \textbf{no existing framework provides real-time, explainable, multi-dimensional orchestration} that:

\begin{enumerate}
    \item Intercepts agent actions \textit{before} execution
    \item Evaluates them across sovereignty, sustainability, compliance, 
          and ethics \textit{simultaneously}
    \item Grounds assessments in \textit{dynamically updated, 
          context-specific} documentation via RAG
    \item Produces \textit{quantitative scores and qualitative 
          explanations} for transparency
    \item Remains \textit{architecturally agnostic} through composition 
          patterns
\end{enumerate}

COMPASS addresses this gap by embedding normative reasoning directly into the operational flow of autonomous agents, transforming ethical governance from a post-hoc audit mechanism into an integral component of agent architecture.

\begin{table}[ht]
\centering
\caption{Comparative analysis of AI governance frameworks}
\label{tab:related_work}
\footnotesize
\setlength{\tabcolsep}{4pt}
\begin{tabular}{@{}lcccccc@{}}
\toprule
\textbf{Framework} & 
\textbf{\shortstack{Real-time\\mediation}} & 
\textbf{\shortstack{Multi-dim.\\evaluation}} & 
\textbf{\shortstack{Dynamic\\grounding}} & 
\textbf{\shortstack{Cultural\\adaptability}} & 
\textbf{\shortstack{Modular\\architecture}} & 
\textbf{\shortstack{Explainable\\decisions}} \\
\midrule
Ethos Institute & $\circ$ & $\bullet$ & $\circ$ & $\circ$ & $\circ$ & $\bullet$ \\
IEEE 7000 & $\circ$ & $\bullet$ & $\circ$ & $\circ$ & $\circ$ & $\bullet$ \\
ETHOS (Web3) & $\circ$ & $\circ$ & $\circ$ & $\circ$ & $\bullet$ & $\bullet$ \\
Ocean Protocol & $\circ$ & $\circ$ & $\circ$ & $\circ$ & $\bullet$ & $\circ$ \\
Constitutional AI & $\bullet$ & $\circ$ & $\circ$ & $\circ$ & $\circ$ & $\circ$ \\
Value-aligned & $\circ$ & $\bullet$ & $\circ$ & $\scriptstyle\bullet$ & $\circ$ & $\bullet$ \\
Generative Agents & $\circ$ & $\circ$ & $\circ$ & $\circ$ & $\circ$ & $\circ$ \\
MT-Bench & $\bullet$ & $\circ$ & $\circ$ & $\circ$ & $\bullet$ & $\scriptstyle\bullet$ \\
\midrule
\textbf{COMPASS} & $\bullet$ & $\bullet$ & $\bullet$ & $\bullet$ & $\bullet$ & $\bullet$ \\
\bottomrule
\end{tabular}
\vspace{2pt}
\begin{center}
{\scriptsize $\bullet$ = Fully supported; $\circ$ = Not supported; $\scriptstyle\bullet$ = Partially supported}
\end{center}
\end{table}

Unlike existing approaches that address governance dimensions in isolation or through post-hoc auditing, COMPASS provides \textit{real-time, explainable orchestration} embedded within agent workflows. By combining RAG-augmented evaluation with an LLM-as-judge methodology across four pillars simultaneously, our framework bridges the gap between policy-level compliance mapping and operational decision-making.

This work provides the methodological foundation and an initial automated evaluation of the agentic framework. Human‑in‑the‑loop validation will be conducted as part of the extended journal version.

The remainder of this paper is organised as follows: Section \ref{Methodology} outlines the proposed methodology, while Section \ref{sec:Result} presents the experimental results. Section \ref{Discussion} provides a detailed discussion and interpretation of the findings. Finally, Section \ref{Conclusion} concludes the study and suggests directions for future research.

\FloatBarrier
\section{Methodology}\label{Methodology}

\subsection{Architectural Overview}
\begin{figure}[H]
    \centering
    \includegraphics[width=1\linewidth]{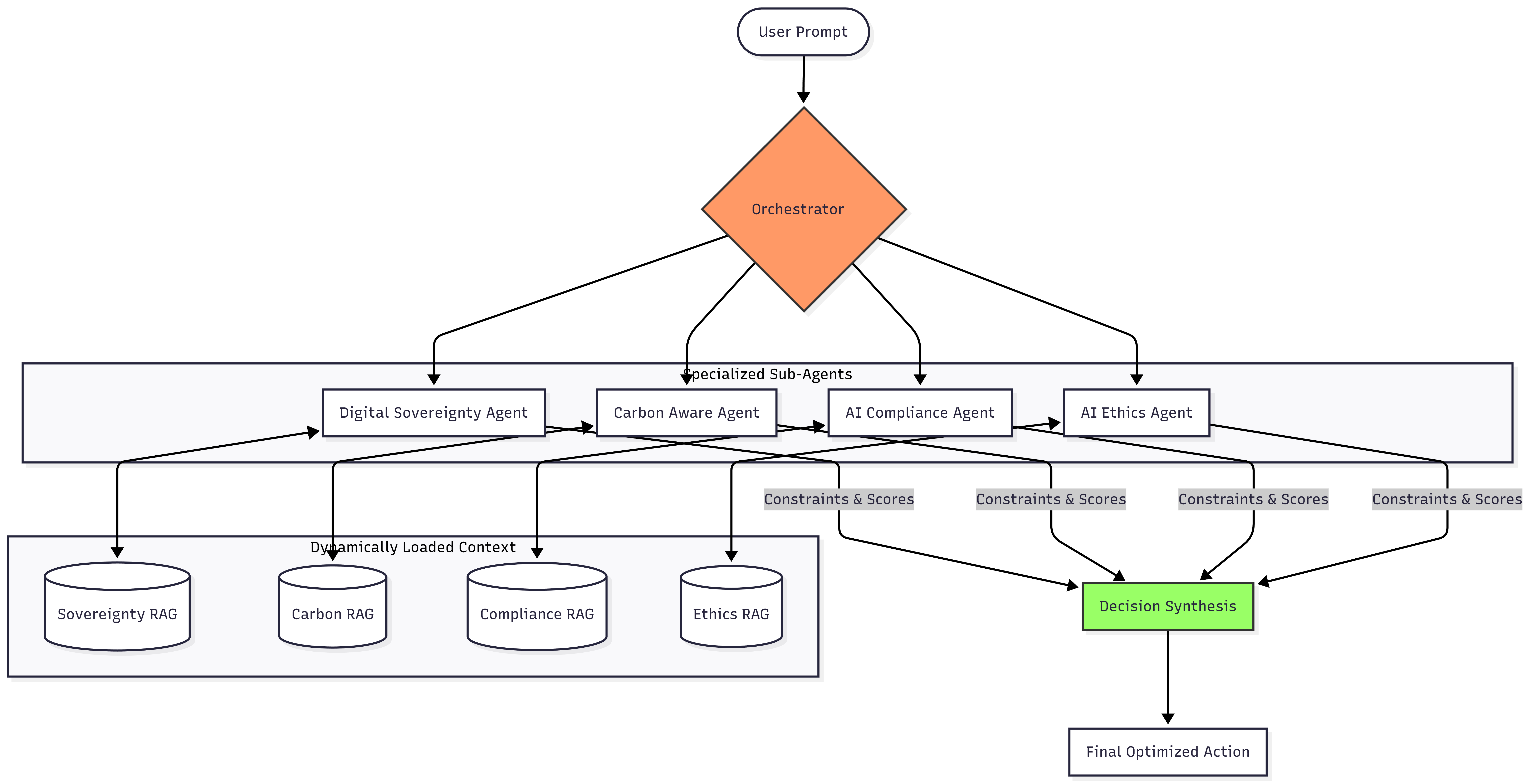}
    \caption{High-level architecture of the multi-agent framework showing the decision-making flow from user input to final optimized action.}
    \label{fig:flowchart}
\end{figure}

Figure \ref{fig:flowchart} illustrates the operational flow of the framework, which mediates between user intent and responsible AI execution. Unlike traditional linear LLM interactions,  introduces a multi-agent orchestration layer.

\textbf{Input Interception:} The process begins when the  Orchestrator receives a user prompt. Instead of generating an immediate response, the orchestrator parses the intent and disseminates the request concurrently to four specialised sub-agents.

\textbf{Contextual Analysis (RAG):} Each sub-agent (Digital Sovereignty, Carbon Awareness, AI Compliance, and AI Ethics) operates independently. They leverage Retrieval-Augmented Generation (RAG) to query their specific dynamic knowledge bases (Vector DBs). For instance, the Compliance Agent retrieves the latest regulatory texts, whilst the Carbon Agent accesses real-time energy intensity data.

\textbf{Synthesis and Conflict Resolution}
The core innovation lies in the Decision Synthesis phase. Once the sub-agents have analysed the request, they return specific constraints and scores rather than raw text.

\textbf{Constraint Aggregation:} The Synthesiser aggregates these inputs, identifying potential conflicts (e.g., a request that is sovereign but computationally distinct and energy-intensive). \textbf{Optimized Output:} Through a weighted scoring mechanism, the framework reconciles these dimensions. The Final Optimised Action is generated only when it satisfies the critical thresholds of all four pillars, ensuring the system’s behaviour remains compliant, ethical, and resource-efficient by design.

\subsection{Structural Design: Composition over Hard-coding}

\begin{figure}[H]
    \centering
    \includegraphics[width=1\linewidth]{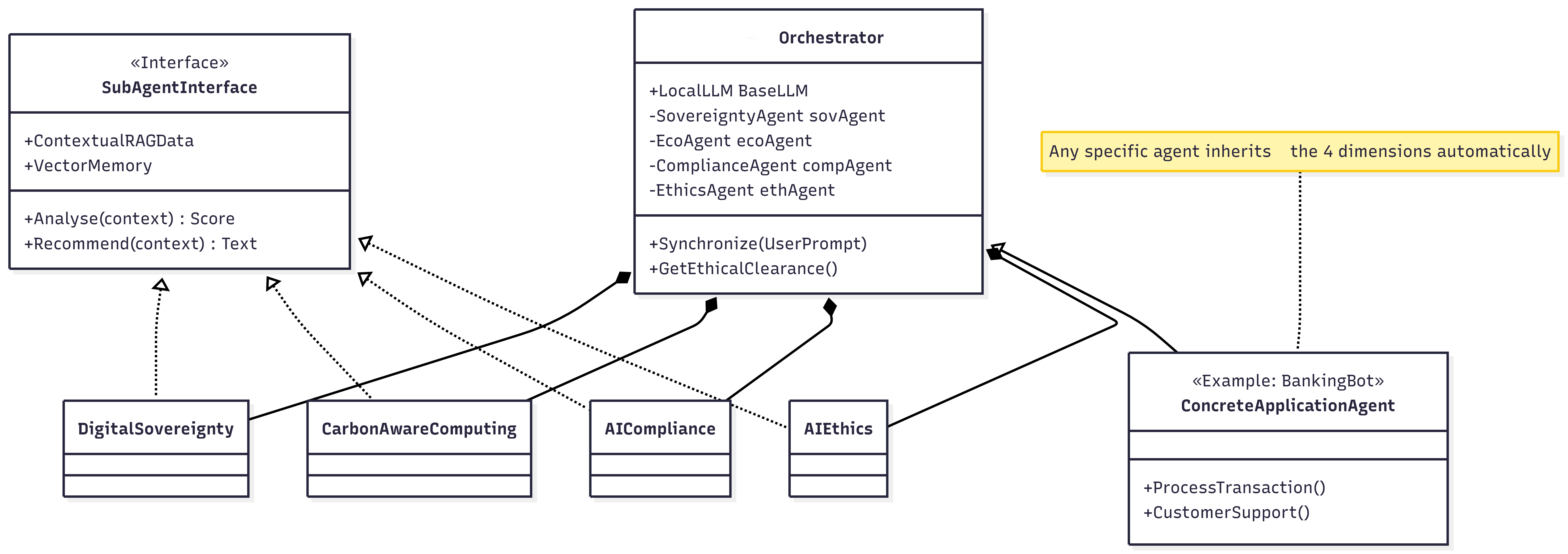}
    \caption{Simplified UML Class Diagram illustrating the object-oriented architecture and extensibility of the  Framework.}
    \label{fig:classes}
\end{figure}

Figure \ref{fig:classes} details the internal structure of the system using standard UML notation. The central component, the Orchestrator, employs a composition pattern (indicated by the solid diamonds) to manage its four strategic dependencies: Sovereignty, Eco, Compliance, and Ethics Agents.
This design choice is critical for software maintenance: it encapsulates each domain of responsibility into a distinct class. It implies that the Orchestrator "owns" these modules, ensuring that no decision-making instance can exist without its constituent ethical engines being instantiated and active.

\textbf{Implementation: Inheritance for "Governance by Design"}

The diagram also demonstrates how developers integrate into real-world scenarios via inheritance (indicated by the open arrow).

\textbf{Extensibility:} The ConcreteApplicationAgent (representing a user-specific application, such as a Banking Bot or HR System) extends the Orchestrator base class.

\textbf{Enforcement:} By virtue of this "Is-A" relationship, the concrete application automatically inherits the Synchronise () and GetEthicalClearance() methods. This enforces an architecture where high-level constraints are embedded in the parent class, making it technically impossible for the application layer to bypass the sovereignty or ethical checks defined in the framework's core.

\subsection{Local LLM and reproductibility}
The system's reasoning and judgement capabilities are driven by an LLM. Its specific configuration was determined empirically and is detailed in Table \ref{tab:LLM_parameters}. 

\begin{table}[h]
    \centering
    \begin{tabular}{ll}
        \toprule
        \textbf{LLM parameters} & \textbf{values} \\
        \midrule
        LLM's name & mistralai\/Mistral-7B-Instruct-v0.2 \\
        \texttt{max\_new\_tokens} & 256 \\
        $temperature$ & 0.7 \\
        $top\_p$ & 0.7 \\
        $repetition\_penalty$ & 1.2 \\
        $do\_sample$ & True \\
        $pad\_token\_id$ & $tokenizer.eos\_token\_id$ \\
        $eos\_token\_id$ & $tokenizer.eos\_token\_id$ \\
        $num\_beams$ & 1 \\
        \botrule
    \end{tabular}
    \caption{Instantiation of prompt keywords with concrete values for a specific query.}
    \label{tab:LLM_parameters}
\end{table}

The specific hyperparameter configuration is detailed in Table \ref{tab:LLM_parameters}. We selected the \textit{Mistral-7B-Instruct-v0.2} model for its high performance-to-size ratio. To ensure the stability and reproducibility required for judging, we adopted a conservative generation strategy. A $temperature$ of $0.7$ controls the randomness of the model’s token sampling, producing outputs that are moderately creative while still maintaining coherence and relevance. A $top\_p$ value of $0.7$ (nucleus sampling) restricts token selection to the smallest set of tokens whose cumulative probability equals $0.7$, ensuring that only relatively probable tokens are considered. Together, these parameters balance diversity and reliability in the generated text.

Furthermore, a repetition penalty of  $1.2$  is applied to prevent loop artefacts, while the output is constrained to a maximum of  $256$  tokens to encourage concise adjudications.

\subsection{Using LLM-as-judge technique}

In automated evaluation, the "LLM-as-a-Judge" paradigm refers to a framework in which a highly capable Large Language Model (e.g., GPT-4) serves as a proxy for human evaluation to assess the quality, safety, or alignment of outputs generated by target models.
Historically, evaluating generative AI has relied either on static $n-gram$ metrics such as BLEU and ROUGE—which often fail to capture semantic nuance—or on human annotation, which is resource-intensive and difficult to scale. The LLM-as-a-Judge approach addresses these limitations by leveraging advanced models' reasoning capabilities to simulate human judgement.
In this configuration, the "Judge" model is provided with a prompt, the system's output, and a specific rubric (e.g., compliance, helpfulness, or reasoning). It then performs a pairwise comparison or assigns a scalar score (e.g., on a Likert scale) to the generated content. Recent studies suggest that, when properly calibrated, these automated judges correlate highly with human agreement, particularly for complex tasks involving reasoning and adherence to specific instructions \cite{zheng2024judging}.

In this study, the LLM-as-a-Judge framework employs both a system prompt and a user prompt for each functionality. For instance, the prompts corresponding to the \textit{Digital Sovereignty} aspect are presented in the listings \ref{lst:system_prompt_sovereignty} and \ref{lst:user_prompt_sovereignty}.\newline

\begin{lstlisting}[caption={Prompt system (Digital Sovereignty example},
                   label={lst:system_prompt_sovereignty},
                   basicstyle=\small\ttfamily,
                   breaklines=true, 
                   breakatwhitespace=true,
                   frame=single]
Context: You are a digital sovereignty expert evaluating technological solutions.

Evaluation Principles:
1. Data Localization: Where is the data stored/processed?
2. Provider Origin: Is the provider based in a jurisdiction compatible with local laws?
3. Technological Control: Is the technology developed/maintained locally?

Task: Evaluate the digital sovereignty compliance of this solution.

Scoring Guide:
- 1.0: Fully compliant (local technology, local hosting, local governance)
- 0.75: Mostly compliant (minor foreign dependencies)
- 0.5: Partially compliant (significant foreign elements but some local control)
- 0.25: Mostly non-compliant (primarily foreign but some local aspects)
- 0.0: Non-compliant (completely foreign-controlled) or insufficient information
- N/A: Not enough information to assign a score.
Note: These values are reference anchors. You are encouraged to use precise intermediate scores (e.g., 0.95, 0.53, 0.20) to reflect the specific nuances of the analysis.

Response format:
ONLY return valid JSON code in this format:
{"score": <float between 0.0 and 1.0, or N/A>, "explanation": "<short text analysis>"}
Examples:
{"score": 0.8, "explanation": "Mistral is a french model, and it is used in France."}
{"score": 0.2, "explanation": "ChatGPT is a american model, and it is used in France."}
{"score": 0.9, "explanation": "Mistral is made in France and also used in France."}
{"score": N/A, "explanation": "Not enough information to assign a score."}
PLEASE VALIDATE TWICE TO BE SURE TO OPEN AND CLOSE THE " AND THE } CORRECTLY.
\end{lstlisting}

\begin{lstlisting}[caption={Prompt user (Digital Sovereignty example},
                   label={lst:user_prompt_sovereignty},
                   basicstyle=\small\ttfamily,
                   breaklines=true, 
                   breakatwhitespace=true,
                   frame=single]
The following would be the Retrieval Augmented Generation data: {RAG1}

This is the request to evaluate:
Test_id: {keyword0}\nCountry: {keyword1}\nGenerative_AI_model: {keyword2}\n
Country_model: {keyword3}\nCountry_data: {keyword4}\nDescription: {keyword5}
Answer in the specified JSON format.
\end{lstlisting}

In this user prompt, the placeholders enclosed in curly braces are dynamically replaced by actual values. Specifically, {RAG1} is substituted with the content retrieved from the reference document via the RAG functionality, as detailed in Subsection \ref{subsec:RAG}. The keywords represent the specific parameters of the query under evaluation. For instance, a data point is defined by the elements listed in Table \ref{tab:keywords}.

\begin{table}[h]
    \centering
    \begin{tabular}{ll}
        \toprule
        \textbf{keywords} & \textbf{values} \\
        \midrule
        keyword0 & "SOV-02" \\
        keyword1 & "Canada"\\
        keyword2 & "Google Gemini" \\
        keyword3 & "France" \\
        keyword4 & "Canada" \\
        keyword5 & "Building a cloud-based AI chatbot for customer support automation" \\
        \botrule
    \end{tabular}
    \caption{Example instantiation of prompt keywords with concrete values for a specific query.}
    \label{tab:keywords}
\end{table}
 
\subsection{RAG}\label{subsec:RAG}
In the context of this method, the information retrieved via RAG enables the customisation of each of the main agent's four functionalities. This adaptation is driven by specific values and the agent's deployment region. Table \ref{tab:RAG} presents examples of reference documents that can serve as sources for the RAG system. 

\begin{table}[h]
    \centering
    \begin{tabular}{ll}
        \toprule
        \textbf{Subjects} & \textbf{Reference documents} \\
        \midrule
        Digital Soverignty & Digital Sovereignty: A Framework to improve  \\   & digital readiness of the Government of Canada, \\ 
        & Government of Canada \cite{secretariat_digital_2025} \\
        Carbon-Aware Computing  &  From Prompts to Power: Measuring the Energy \\ 
         & Footprint of LLM Inference \cite{caravaca2025promptspowermeasuringenergy} \\
        AI Compliance & Artificial Intelligence Act, European Union. \cite{noauthor_act_nodate} \\
        AI Ethics & Montreal Declaration for Responsible AI. \cite{montreal2018declaration} \\
        \botrule
    \end{tabular}
    \caption{Reference documents used for the tests. These may vary by country and users' preferences.}
    \label{tab:RAG}
\end{table}

The listings \ref{lst:rag_system_prompt_sovereignty} and \ref{lst:rag_user_prompt_sovereignty} present the System and User prompts for the agent's \textit{Digital Sovereignty} aspect.\newline

\begin{lstlisting}[caption={System prompt for RAG-based sovereignty evaluation},
                   label={lst:rag_system_prompt_sovereignty},
                   basicstyle=\small\ttfamily,
                   breaklines=true, 
                   breakatwhitespace=true,
                   frame=single]
You must extract the most valuable information in this document.
\end{lstlisting}

\begin{lstlisting}[caption={Prompt User for RAG (Digital Sovereignty example)},
                   label={lst:rag_user_prompt_sovereignty},
                   basicstyle=\small\ttfamily,
                   breaklines=true, 
                   breakatwhitespace=true,
                   frame=single]
What are the document's key points?
\end{lstlisting}

The main agent supports instantiation with an optional setting that activates RAG features across all four elements. In the absence of RAG support, the system functions but with reduced efficacy, requiring the LLM-as-judge to evaluate context unaided. Conversely, when the model is augmented via RAG, the judgement process becomes customised to user-defined values, leading to the performance improvements discussed in Section \ref{sec:Result}. 

\subsection{Metrics}


BERTScore provides a semantically informed alternative to traditional n‑gram‑based evaluation metrics by comparing the contextual embeddings of candidate and reference texts rather than their surface lexical overlap. Because it leverages pretrained transformer representations, BERTScore captures fine‑grained semantic similarity, making it substantially more aligned with human judgements than metrics such as BLEU or ROUGE, especially in tasks where paraphrasing and lexical variation are common. Its use has become increasingly widespread in the assessment of generative models, as it preserves sensitivity to meaning while remaining robust to vocabulary choice and word-order differences.\newline

Equation 1 (Contextual Embeddings). 
Both the reference and candidate sentences are mapped to contextual embeddings that capture semantic meaning informed by surrounding words. 
These representations are generated by large pretrained language models such as BERT, RoBERTa, XLNet, and XLM.

\[
R_{\text{BERT}} = 
\frac{1}{|x|}
\sum_{x_i \in x}
\max_{\hat{x}_j \in \hat{x}}
\mathbf{x}_i^{\top} \hat{\mathbf{x}}_j
\tag{1}
\]

Equation 2 (Cosine Similarity).  
Semantic similarity between sentences is quantified using cosine similarity computed over contextual embeddings, enabling robust matching even when the lexical wording differs.

\[
P_{\text{BERT}} =
\frac{1}{|\hat{x}|}
\sum_{\hat{x}_j \in \hat{x}}
\max_{x_i \in x}
\mathbf{x}_i^{\top} \hat{\mathbf{x}}_j
\tag{2}
\]

Equation 3 (Token Matching for Precision and Recall).  
Token‑level alignment is performed in both directions: each candidate token is matched to the most similar reference token, and each reference token is matched to the most similar candidate token.  
These directional similarities yield recall and precision, which are combined via their harmonic mean to produce the BERTScore F1.

\[
F_{\text{BERT}} =
2 \cdot
\frac{P_{\text{BERT}} \cdot R_{\text{BERT}}}
{P_{\text{BERT}} + R_{\text{BERT}}}
\tag{3}
\]

Beyond these core computations, the contribution of each token may be modulated using Inverse Document Frequency (IDF) weighting.  
This optional refinement emphasises rare or domain‑specific words, thereby better aligning the scoring process with the semantic distribution of specialised corpora.

To improve interpretability and ensure comparability across datasets, BERTScore values can also be rescaled using baseline statistics derived from large monolingual corpora such as Common Crawl.  
This linear transformation adjusts scores to fall within a more intuitive and empirically grounded range.

\[
\hat{R}_{\text{BERT}} =
\frac{R_{\text{BERT}} - b}{1 - b}
\]

In the present work, BERTScore is used to compare the differences between the LLM‑as‑judge with and without RAG augmentation. Although human‑in‑the‑loop validation will be introduced at a later stage, this evaluation already illustrates the degree of personalisation enabled by RAG‑based data augmentation relative to the user-provided document.

\section{Results}\label{sec:Result}

Experimental results were generated for the agent's four core capabilities. In this initial iteration of the framework, only the evaluation mechanism has been implemented for each sub-agent; however, each domain holds significant potential for future expansion and the integration of additional features. These evaluation capabilities—covering \textit{Digital Sovereignty}, \textit{Carbon-Aware Computing}, \textit{AI Compliance}, and \textit{AI Ethics}—serve as the foundation of our method. They demonstrate the core premise: that agents inheriting from the  framework will possess the requisite intelligence and domain knowledge to enhance the acceptability of agentic AI. Tables \ref{tab:ResultsScores_SOV}, \ref{tab:ResultsScores_CAR}, \ref{tab:ResultsScores_COM}, and \ref{tab:ResultsScores_ETH} present the results across these dimensions.

The \textit{Test id.} column indicates the unique test identifier and its category: \textit{SOV} for \textit{Digital Sovereignty}, \textit{CAR} for \textit{Carbon-Aware Computing}, \textit{COM} for \textit{AI Compliance}, and \textit{ETH} for \textit{AI Ethics}. The \textit{Score without RAG} and \textit{Score with RAG} columns display the ratings assigned by the \textit{LLM-as-a-judge}. The  $\Delta Score$  denotes the variance between these ratings, highlighting the impact of the RAG-retrieved information. Finally, the \textit{Similarity} field provides the \textit{BERTScore}, measuring the semantic similarity between the explanatory texts generated by the LLM to justify its judgments.


\begin{table}[h]
\begin{tabular}{lrrrr}
\toprule
\textbf{Test id.} & \textbf{Score without RAG} & \textbf{Score with RAG} & \textbf{$\Delta$ Score} & \textbf{Similarity} \\
\midrule
SOV-01 & 0.25 & 0.50 & +0.25 & 74.7\% \\
SOV-02 & 0.25 & 0.25 & +0.00 & 76.1\% \\
SOV-03 & 0.25 & 0.25 & +0.00 & 77.0\% \\
SOV-04 & 0.25 & 0.25 & +0.00 & 79.2\% \\
SOV-05 & 0.50 & 0.50 & +0.00 & 80.4\% \\
SOV-06 & 0.25 & 0.50 & +0.25 & 80.3\% \\
SOV-07 & 0.25 & 0.50 & +0.25 & 77.6\% \\
SOV-08 & 0.50 & 0.75 & +0.25 & 78.0\% \\
SOV-09 & 0.25 & 0.25 & +0.00 & 80.7\% \\
SOV-10 & 0.25 & 0.50 & +0.25 & 75.5\% \\
\bottomrule
\end{tabular}
\caption{Results for the evaluation of the sovereignty aspect.}
\label{tab:ResultsScores_SOV}
\end{table}

\begin{table}[h]
\begin{tabular}{lrrrr}
\toprule
\textbf{Test id.} & \textbf{Score without RAG} & \textbf{Score with RAG} & \textbf{$\Delta$ Score} & \textbf{Similarity} \\
\midrule
CAR-01 & 0.85 & 0.80 & -0.05 & 75.8\% \\
CAR-02 & 0.50 & 0.50 & +0.00 & 81.4\% \\
CAR-03 & 0.50 & 0.50 & +0.00 & 85.3\% \\
CAR-04 & 0.65 & 0.60 & -0.05 & 79.6\% \\
CAR-05 & 0.50 & 0.50 & +0.00 & 74.3\% \\
CAR-06 & 0.50 & 0.53 & +0.03 & 82.7\% \\
CAR-07 & 0.50 & 0.55 & +0.05 & 79.2\% \\
CAR-08 & 0.50 & 0.50 & +0.00 & 88.7\% \\
CAR-09 & 0.75 & 0.75 & +0.00 & 76.1\% \\
CAR-10 & 0.50 & 0.50 & +0.00 & 80.0\% \\
\bottomrule
\end{tabular}
\caption{Results for the evaluation of the carbon aspect.}
\label{tab:ResultsScores_CAR}
\end{table}

\begin{table}[h]
\begin{tabular}{lrrrr}
\toprule
\textbf{Test id.} & \textbf{Score without RAG} & \textbf{Score with RAG} & \textbf{$\Delta$ Score} & \textbf{Similarity} \\
\midrule
COM-01 & 0.50 & 0.25 & -0.25 & 76.4\% \\
COM-02 & 0.50 & 0.25 & -0.25 & 75.6\% \\
COM-03 & 0.50 & 0.50 & +0.00 & 75.5\% \\
COM-04 & 0.50 & 0.50 & +0.00 & 76.8\% \\
COM-05 & 0.50 & 0.25 & -0.25 & 77.4\% \\
COM-06 & N/A & 0.50 & N/A & 67.9\% \\
COM-07 & 0.50 & 0.25 & -0.25 & 70.2\% \\
COM-08 & 0.50 & N/A & N/A & 67.0\% \\
COM-09 & 0.50 & N/A & N/A & 66.2\% \\
COM-10 & 0.50 & 0.25 & -0.25 & 78.6\% \\
\bottomrule
\end{tabular}
\caption{Results for the evaluation of the compliance aspect.}
\label{tab:ResultsScores_COM}
\end{table}

\begin{table}[h]
\begin{tabular}{lrrrr}
\toprule
\textbf{Test id.} & \textbf{Score without RAG} & \textbf{Score with RAG} & \textbf{$\Delta$ Score} & \textbf{Similarity} \\
\midrule
ETH-01 & 0.75 & 0.75 & +0.00 & 76.4\% \\
ETH-02 & 1.00 & 0.95 & -0.05 & 81.4\% \\
ETH-03 & 0.50 & 0.25 & -0.25 & 77.2\% \\
ETH-04 & 0.50 & 0.50 & +0.00 & 78.2\% \\
ETH-05 & 0.50 & 0.50 & +0.00 & 90.8\% \\
ETH-06 & 0.50 & 0.50 & +0.00 & 77.9\% \\
ETH-07 & 0.50 & 0.50 & +0.00 & 72.9\% \\
ETH-08 & 0.25 & 0.25 & +0.00 & 79.2\% \\
ETH-09 & 0.50 & 0.50 & +0.00 & 79.6\% \\
ETH-10 & 0.50 & 0.50 & +0.00 & 85.5\% \\
\bottomrule
\end{tabular}
\caption{Results for the evaluation of the ethical aspect.}
\label{tab:ResultsScores_ETH}
\end{table}

\begin{figure}[h]
    \centering
    \includegraphics[width=0.50\linewidth]{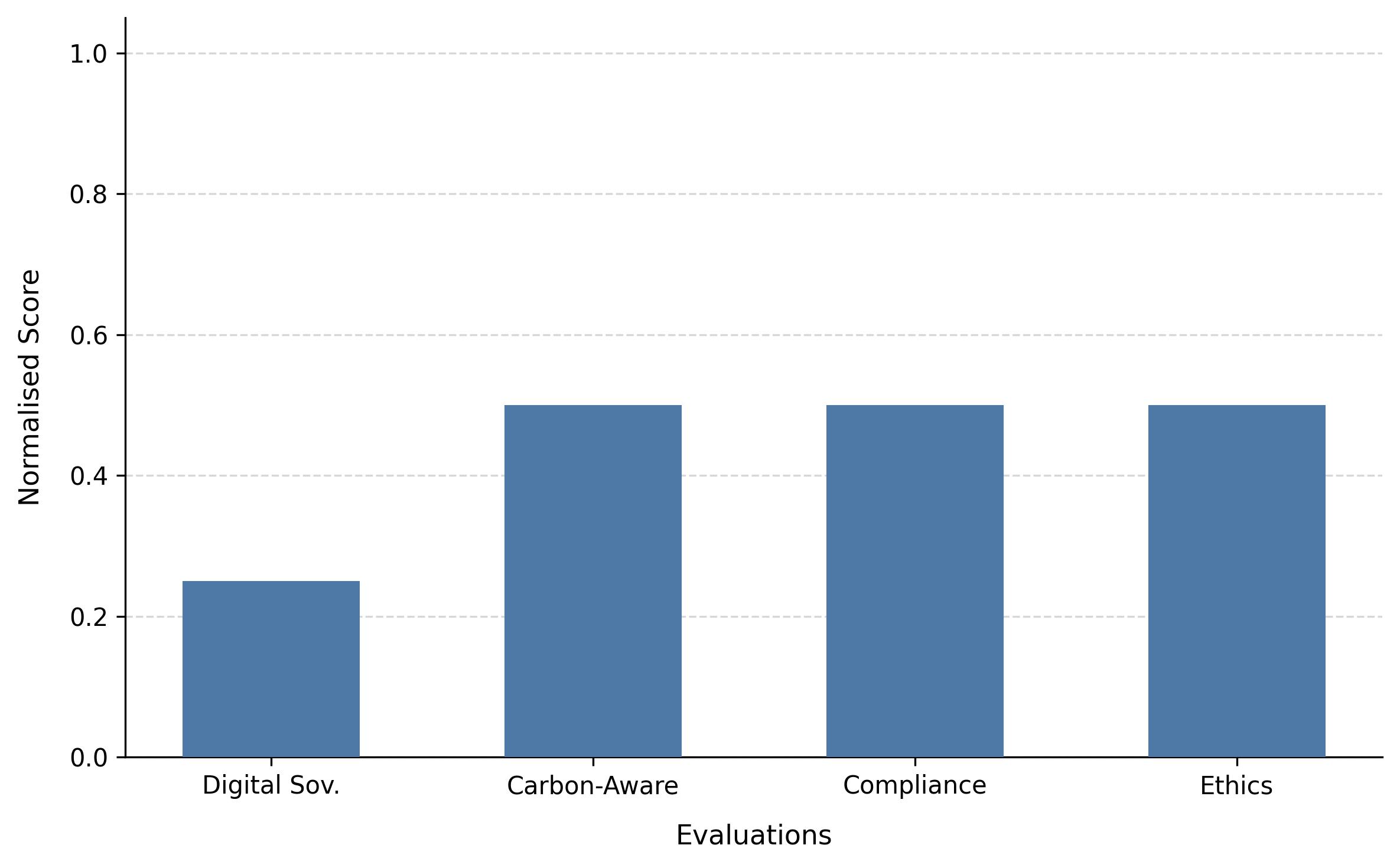}
    \caption{Explainability: Bar chart for use case (test.id): SOV-05, CAR-05, COM-05, ETH-05.}
    \label{fig:bar_chart}
\end{figure}

\begin{figure}[h]
    \centering
    \includegraphics[width=0.50\linewidth]{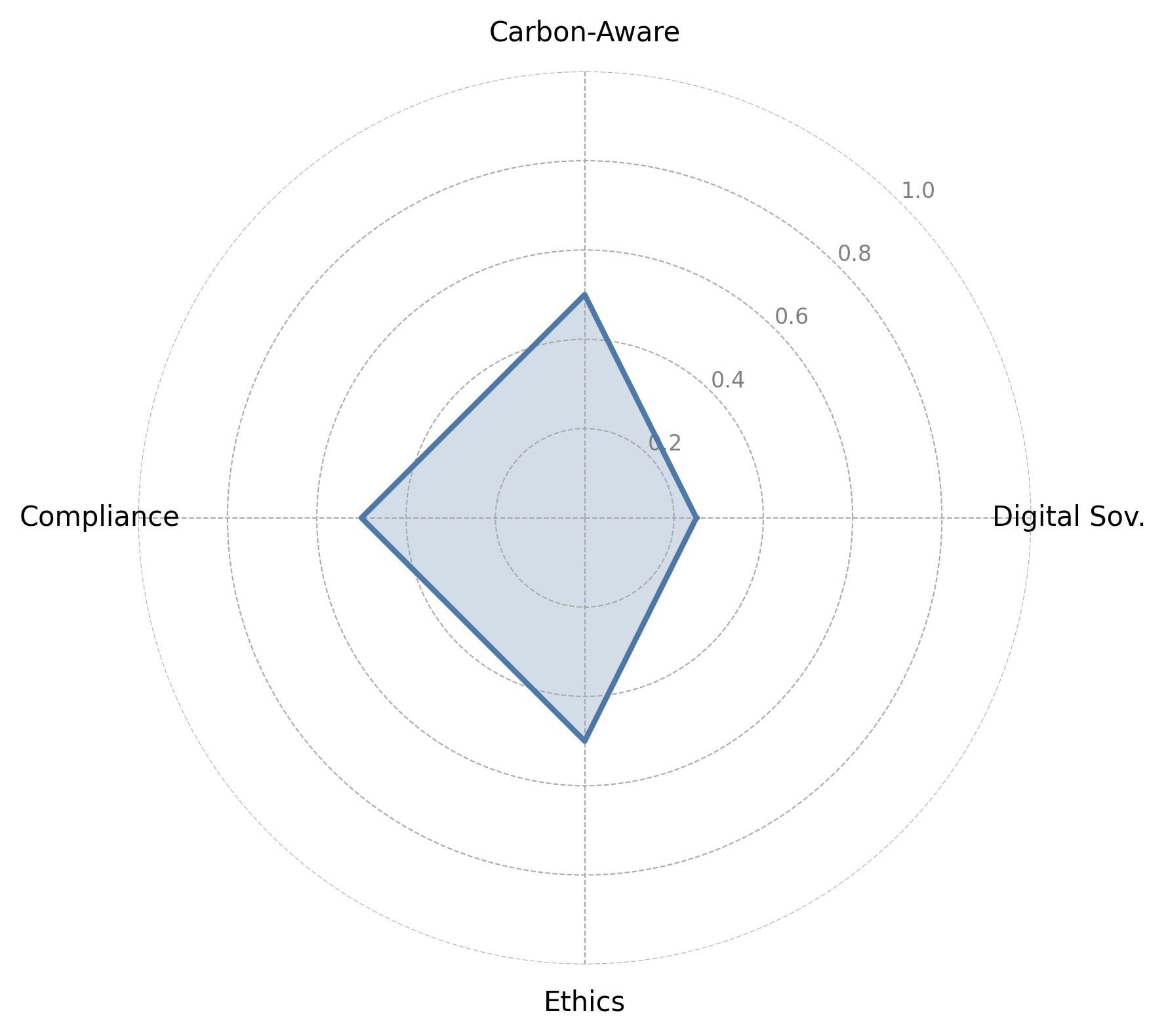}
    \caption{Explainability: radar graphic for use case (test.id): SOV-05, CAR-05, COM-05, ETH-05.}
    \label{fig:radar_chart}
\end{figure}

As illustrated in Fig. \ref{fig:flowchart}, each sub-agent transmits a set of constraints (information) alongside a corresponding score to the decision synthesis module (depicted as the green box in the figure). Presently, this module consolidates the received data and presents it as a bar chart and a radar plot, thereby summarising the entire process. Following their utilisation by this explainability module, the generated text and score serve as the basis for determining the optimal course of action, if applicable. This final step, though already conceptualised, represents the logical culmination of our proposed framework. Its implementation is reserved for the next iteration of our methodology and will be addressed in future work. The two generated visualisations, depicted in Fig. \ref{fig:bar_chart} and Fig. \ref{fig:radar_chart}, provide an explanatory representation of the decision synthesis process.

\FloatBarrier
\section{Discussion}\label{Discussion}

\subsection{Comparison with Concurrent Work}

While the blockchain-based ETHOS framework \cite{ethos_blockchain_2024} provides robust post-hoc accountability through immutable audit trails, COMPASS operates at a fundamentally different temporal point in the agent lifecycle. ETHOS excels at \textit{verifying what happened}, whereas COMPASS focuses on \textit{guiding what should happen}. These approaches are complementary: COMPASS decisions could be logged to an ETHOS registry, creating an end-to-end governance pipeline from real-time mediation to permanent record-keeping.

Similarly, meta-frameworks like Ethos Institute's harmonisation system \cite{ethos_institute} provide invaluable mappings between regulatory requirements. COMPASS operationalises these mappings by translating abstract compliance principles into executable evaluation logic grounded in specific documentation via RAG. Where meta-frameworks answer "what regulations apply?", COMPASS answers "does this action comply?"

\begin{table}[ht]
    \centering
    \small 
    \caption{Qualitative comparison of the  Framework against existing AI paradigms.}
    \label{tab:comparison_matrix}
    
    \begin{tabularx}{\linewidth}{l >{\centering\arraybackslash}X >{\centering\arraybackslash}X >{\centering\arraybackslash}X >{\centering\arraybackslash}X}
        \toprule
        \textbf{Feature} & \textbf{Vanilla LLM} & \textbf{Standard RAG} & \textbf{sLLM Agents} & \textbf{ (Ours)} \\ 
        \midrule
        
        \textbf{Reasoning} & Black-box & Retrieval-based & Goal-oriented & \textbf{Value-aligned} \\
        \addlinespace
        
        \textbf{Adaptability} & No & Yes & Partial & \textbf{Yes} \\
        \textit{\scriptsize{(e.g. to new laws)}} & & & & \\
        \addlinespace
        
        \textbf{Sovereignty} & Low & Medium & Medium & \textbf{High} \\
        \textit{\scriptsize{(Local exec.)}} & & & & \\
        \addlinespace
        
        \textbf{Multi-dim. Check} & No & No & No & \textbf{Yes} \\
        \textit{\scriptsize{(Conflict solver)}} & & & & \\
        \addlinespace
        
        \textbf{Explainability} & Parametric & Retrieved Context & Action Traces & \textbf{Context \& Adjudication} \\
        
        \bottomrule
    \end{tabularx}
    
    \vspace{0.1cm}
    \footnotesize{\textit{Note: "sLLM Agents" refers to standard task-oriented frameworks like AutoGPT.}}
\end{table}

To validate the architectural necessity of the retrieval mechanism, we conducted an ablation study comparing the agent's performance in its \textit{Vanilla} state (an LLM reliant solely on pretraining weights) versus its \textit{Augmented} state (a Framework with RAG). The results highlight a critical distinction in the reliability of the agentic decisions.
While the quantitative scores ( $\delta Score$ ) may occasionally converge—indicating that the base model possesses a foundational understanding of high-level ethical or sovereign concepts—the qualitative analysis reveals the limits of the non-augmented approach. Without access to external references, the standard LLM tends to generate generic justifications or, in worst-case scenarios, hallucinations regarding specific local regulations (e.g., hallucinating a non-existent Canadian privacy statute). Conversely, integrating RAG ensures that the agent's reasoning is grounded in verified, up-to-date documents. The semantic similarity metrics and the explicit citations in the generated explanations evidence this grounding. Consequently, the framework shifts the agent's behaviour from probabilistic guessing to evidence-based adjudication, a requisite feature for social acceptability in sensitive deployment environments.

While the present study relies exclusively on automated evaluation using an LLM-as-a-judge, it does not yet include human‑in‑the‑loop validation, which remains essential for establishing empirical reliability beyond model‑to‑model agreement. Future work will incorporate a structured human evaluation phase to assess inter‑rater consistency and to benchmark the judge’s decisions against expert and non‑expert annotators. In parallel, we plan to investigate whether performance and robustness can be improved through a mixture‑agent approach, in which complementary agents specialise in distinct assessment dimensions and collectively provide a more robust evaluation signal.

Naturally, conflicts may arise between sub-agents, as one may assign a favourable score while another attributes an unfavourable one. For instance, addressing concerns about local data centres and LLMs could inadvertently reduce energy efficiency and increase GHG emissions. However, such scenarios involve numerous subtleties that require careful consideration. It is the role of the orchestrator to elucidate these nuances, a task that would prove considerably more challenging without our  Agent, which encapsulates all core functionalities.

While this method demonstrates potential, its capacity to select optimal actions remains underdeveloped. At present, it can generate explanations and assign scores, but future iterations could actively enforce compliance with our ethical framework. For example, a practical application might involve intelligently selecting an appropriate LLM for a given task—one that minimises bias, operates locally, and, where feasible, consumes less energy.
Another concrete action could be issuing real-time alerts to users, such as flagging when a task violates local regulations.
Moving forward, innovative solutions will be essential to fully realise this system’s potential.\newline

\noindent Several key avenues for improvement warrant exploration in future work:

\begin{enumerate}

    \item This foundational method establishes the groundwork for the further development of the four sub-agents. Multiple features specific to each sub-agent can be added to refine and enhance the method. It is intended to serve as the core framework, with each functionality to be further developed and elaborated.
    
    \item Document Curation for RAG – A systematic study is needed to identify and select the most relevant documents for RAG to ensure high-quality inputs. Additionally, expanding the system to support a broader corpus of reference documents could further enhance performance.
    
    \item LLM-as-Judge Refinement – The evaluation capabilities of LLM-based judging mechanisms require improvement, particularly in mitigating inherent biases that may influence decision-making.
    
    \item Human-in-the-Loop Validation – Current limitations include a lack of structured human oversight. Future research should investigate hybrid validation methods that integrate expert review to refine and verify the system’s outputs.
    
    \item Code-Level Ethical Analysis – Beyond data analysis, the Agent will eventually extend its evaluation to the underlying code of augmented agents, enabling a more comprehensive and granular ethical assessment.
\end{enumerate}

\section{Conclusion}\label{Conclusion}

This paper introduces the COMPASS Framework, a novel multi-agent orchestration architecture designed to address the critical challenge of aligning agentic AI systems with contemporary imperatives of digital sovereignty, environmental sustainability, regulatory compliance, and ethical responsibility. Unlike conventional LLM-based approaches that prioritise task performance in isolation, our framework systematically integrates value-aligned reasoning into the operational logic of autonomous agents through a modular, extensible design.
The proposed system comprises an Orchestrator that coordinates four specialised sub-agents, each responsible for evaluating a distinct normative dimension. By leveraging RAG to ground assessments in verified, context-specific documentation, the framework mitigates the hallucination risks inherent to purely parametric models while enabling adaptability across diverse regulatory and cultural contexts. Our LLM-as-a-judge methodology provides both quantitative scoring and qualitative explanations, facilitating transparency and interpretability. These features constitute essential prerequisites for the social acceptability of AI systems in sensitive deployment environments.
Automated evaluation demonstrates that RAG integration substantially enhances semantic coherence, as measured by BERTScore, and ensures that agent reasoning is evidence-based rather than speculative. The architecture's composition-based design permits seamless integration into existing agentic workflows, thereby extending responsible AI practices across application domains without requiring fundamental restructuring of host systems.
Nevertheless, several limitations warrant acknowledgement. Firstly, the current study relies exclusively on automated evaluation; human-in-the-loop validation remains essential to empirically assess inter-rater reliability and benchmark model judgements against expert consensus. Secondly, whilst the framework successfully generates scores and explanations, its capacity to autonomously select and execute corrective actions remains underdeveloped. Future iterations must implement decision-making modules capable of dynamically selecting optimal LLMs based on sovereignty constraints or issuing real-time alerts when tasks violate regulatory boundaries. Thirdly, potential conflicts between sub-agents require sophisticated orchestration strategies. For instance, tensions between local data sovereignty and carbon efficiency necessitate a transparent resolution of trade-offs.
Looking forward, several research avenues merit exploration. Document curation methodologies for RAG must be systematically investigated to optimise retrieval quality and expand reference corpora. The LLM-as-judge mechanism itself warrants refinement, particularly in terms of bias mitigation and consistency across diverse evaluation contexts. A mixture-of-agents approach, in which complementary models provide robust assessment signals, may further enhance robustness. Finally, extending the framework's analytical scope to code-level ethical auditing would facilitate more comprehensive governance. This extension would enable examination of the underlying algorithms of augmented agents.
In conclusion, the  Framework represents a foundational step towards reconciling the operational autonomy of agentic AI with the normative constraints essential for responsible deployment. By embedding multi-dimensional ethical reasoning directly into agent architectures, this work contributes to the broader endeavour of ensuring that advanced AI systems remain aligned with human values, societal norms, and planetary boundaries. As agentic AI continues to permeate critical domains, from healthcare to public administration, frameworks such as  will prove indispensable in fostering both innovation and trustworthiness.

\section{Statements and Declarations}%
\label{sec:Statement}

\noindent\textbf{Fundings}
This research received no specific grant from any funding agency in the public, commercial, or not-for-profit sectors.\newline

\noindent\textbf{Conflict of interest} 
The authors confirm there are no conflicts of interest. \newline

\noindent\textbf{Ethical approval}
The work uses publicly available and non-identifiable information. No ethical approval was needed.\newline

\noindent\textbf{Consent to participate} 
Not applicable, as no human participants were involved in the evaluation of our study. \newline

\noindent\textbf{Consent for publication} 
Consent for publication is not applicable as this study does not contain any identifiable data.\newline

\noindent\textbf{Availability of data and material}
Not available at this time. All data and materials will be made available upon publication in a journal. \newline

\noindent\textbf{Code availability}
Not available at this time. All code will be made available upon publication in a journal.\newline 

\noindent\textbf{Authors' contributions}
JSD: Conceptualisation, Methodology, Validation, Resources, Writing - Review \& Editing, Supervision, Project administration. 

This manuscript benefited from the assistance of an AI-based language model (LLMs, Grammarly, DeepL) for improving the clarity and fluency of academic English. All content was reviewed and approved by the authors.\newline

\bibliography{sn-bibliography}

\end{document}